\documentclass[10pt,letterpaper]{article}

\usepackage{cogsci}
\usepackage{pslatex}
\usepackage{apacite}

\usepackage{amsmath}
\usepackage{amssymb}

\usepackage{pdfpages}
\usepackage{todonotes}

\title{A computational investigation of sources of variability\\
 in sentence comprehension difficulty in aphasia}
 
\author{{\large\bf Paul M\"{a}tzig (pmaetzig@uni-potsdam.de)} \\
        University of Potsdam, Human Sciences Faculty, Department Linguistics,\\
        24--25 Karl-Liebknecht-Str., Potsdam 14476, Germany\\
        \AND {\large\bf Shravan Vasishth, (vasishth@uni-potsdam.de)} \\
        University of Potsdam, Human Sciences Faculty, Department Linguistics,\\
        24--25 Karl-Liebknecht-Str., Potsdam 14476, Germany\\
        \AND {\large\bf Felix Engelmann (felix.engelmann@manchester.ac.uk)} \\
        The University of Manchester, School of Health Sciences\\
        Child Study Centre, Coupland 1, Oxford Road, Manchester M13 9PL\\
        \AND {\large\bf David Caplan (dcaplan@partners.org)} \\
        Massachusetts General Hospital \\
        175 Cambridge St, \#340, Boston, Massachusetts 02114}

\begin{document}

\maketitle

\begin{abstract} We present a computational evaluation of three hypotheses about sources of deficit in sentence comprehension in aphasia: slowed processing, intermittent deficiency, and resource reduction.  The ACT-R based \citeA{LewisVasishth2005} model is used to implement these three proposals. Slowed processing is implemented as slowed default production-rule firing time; intermittent deficiency as increased random noise in activation of chunks in memory; and resource reduction as reduced goal activation. As data, we considered subject vs.\ object relatives whose matrix clause contained either an NP or a reflexive, presented in a self-paced listening modality to 56 individuals with aphasia (IWA) and 46 matched controls. The participants heard the sentences and carried out a picture verification task to decide on an interpretation of the sentence. These response accuracies are used to identify the best parameters (for each participant) that correspond to the three hypotheses mentioned above. We show that controls have more tightly clustered (less variable) parameter values than IWA; specifically, compared to controls, among IWA there are more individuals with low goal activations, high noise, and slow default action times. This suggests that (i) individual patients show differential amounts of deficit along the three dimensions of slowed processing, intermittent deficient, and resource reduction, (ii) overall, there is evidence for all three sources of deficit playing a role, and (iii) IWA have a more variable range of parameter values than controls. In sum, this study contributes a proof of concept of a quantitative implementation of, and evidence for, these three accounts of comprehension deficits in aphasia.

\end{abstract}

\section{Introduction}

In healthy adults, 
sentence comprehension has long been argued to be influenced by individual differences; a commonly assumed source is differences in working memory capacity  \cite{dc80,jc92}. 
Other factors such as age \cite{CaplanWaters2005} and cognitive control \cite{novick2005cognitive} have also been implicated.

An important question that has not received much attention in the computational psycholinguistics literature is: what are sources of individual differences in healthy adults versus impaired populations, such as individuals with aphasia (IWA)? 
It is well-known that
sentence processing performance in IWA is characterised by a performance deficit that expresses itself as slower overall processing times, and lower accuracy in question-response tasks (see literature review in \citeNP{PatilEtAl2016}). 
These performance deficits are especially pronounced when IWA have to engage with sentences that have non-canonical word order and that are semantically reversible, e.g.\ Object-Verb-Subject versus Subject-Verb-Object sentences \cite{hanneetal11}. 

Regarding the underlying nature of this deficit in IWA, there is a consensus that some kind of disruption is occurring in the syntactic comprehension system. The exact nature of this disruption, however, is not clear. 
Although a broad range of proposals exist (see \citeNP{PatilEtAl2016}), 
we focus on three influential proposals here:
\begin{enumerate}
\item \textit{Intermittent deficiencies}:
\citeA{CaplanEtAl2015} suggest that occasional temporal breakdowns of parsing mechanisms capture the observed behaviour. 
\item \textit{Resource reduction}:
A third hypothesis, due to \citeA{Caplan2012},  is that the deficit is caused by a reduction in resources related to sentence comprehension. 
\item \textit{Slowed processing}:
\citeA{BurkhardtEtAl2003} argue that a slowdown in parsing mechanisms can best explain the processing deficit.  
\end{enumerate} 

Computational modelling can help evaluate these different proposals quantitatively. 
Specifically, the cue-based retrieval account of \citeA{LewisVasishth2005}, which was developed within the ACT-R framework \cite{AndersonEtAl2004}, is a computationally implemented model of unimpaired sentence comprehension 
 that has been used to model a broad array of empirical phenomena in sentence processing relating to similarity-based interference effects 
\cite{LewisVasishth2005,NicenboimVasishth2017StanCon,VBLD07,EngelmannJaegerVasishthSubmitted2016} and the interaction between oculomotor control and  sentence comprehension \cite{Engelmanna}.\footnote{The model can be downloaded in its current form from 
https://github.com/felixengelmann/act-r-sentence-parser-em.}

The \citeA{LewisVasishth2005} model is particularly attractive for studying sentence comprehension because it relies on the general constraints on cognitive processes that have been laid out in the ACT-R framework. This makes it possible to investigate whether sentence processing could be seen as being subject to the same general cognitive constraints as  any other information processing task, which does not entail that there are no language specific constraints on sentence comprehension.
A further advantage of the \citeA{LewisVasishth2005} model in the context of theories of processing deficits in aphasia is that several of its numerical parameters (which are part of the general ACT-R framework) can be interpreted as  implementing the three proposals mentioned above.

In \citeA{PatilEtAl2016}, the \citeA{LewisVasishth2005} architecture was used to model aphasic sentence processing on a small scale, using data from seven patients. 
They modelled proportions of fixations in a visual world task, response accuracies and response times for empirical data of a sentence-picture matching experiment by \citeA{hanneetal11}. Their goal was to test two of the three hypotheses of sentence comprehension deficits mentioned above, slowed processing and intermittent deficiency.  

In the present work, we provide a proof of concept study that goes beyond \citeA{PatilEtAl2016} by evaluating the evidence for the three hypotheses---slowed processing, intermittent deficiencies, and resource reduction---using a larger data-set from \citeA{CaplanEtAl2015} with 56 IWA and 46 matched controls.

Before we describe the modelling carried out in the present paper and the data used for the evaluation, 
we first introduce the cognitive constraints assumed in the \citeA{LewisVasishth2005} model that are relevant for this work, and show how the theoretical approaches to the aphasic processing deficit can be implemented using specific model parameters. 
Having introduced the essential elements of the model architecture, 
we simulate comprehension question-response accuracies for unimpaired controls and IWA, and then fit the simulated accuracy data to published data
\cite{CaplanEtAl2015} from controls and IWA. When fitting individual participants, we vary three parameters that map to the three theoretical proposals mentioned above. The goal was to determine whether the distributions of parameter values furnish any support for any of the three sources of deficits in processing. 
We expect that if there is a tendency in one parameter to show non-default values in IWA, for example slowed processing,
then there is support for the claim that slowed processing is an underlying source of processing difficulty in IWA. Similar predictions hold for the other two constructs, intermittent deficiency and resource reduction; and for combinations of the three proposals.

\section{Constraints on sentence comprehension in the \protect\citeA{LewisVasishth2005} model}

In this section, we describe some of the constraints assumed in the \citeA{LewisVasishth2005} sentence processing model. Then, we discuss the model parameters that can be mapped to the three theoretical proposals for the underlying processing deficit in IWA.

The ACT-R architecture assumes a distinction between long-term declarative memory and procedural knowledge. The latter is implemented as a set of rules, consisting of condition-action pairs known as production rules. These production rules operate on units of information known as chunks, which are elements in declarative memory that are defined in terms of feature-value specifications. For example, a noun like \textit{book} could be stored as a feature-value matrix that states that the part-of-speech is nominal, number is singular, and animacy status is inanimate:

\begin{equation*}
\begin{pmatrix}
  \text{pos} & \mathit{nominal} \\
  \text{number} & \mathit{sing} \\
  \text{animate} & \mathit{no}
\end{pmatrix}
\end{equation*}

 Each chunk is associated an \emph{activation}, a numeric value that determines the probability and latency of access from declarative memory. Accessing chunks in declarative memory happens via a cue-based retrieval mechanism. For example, if the noun \textit{book} is to be retrieved, cues such as \{part-of-speech nominal, number singular, and animate no\} could be used to retrieve it.  Production rules are written to trigger such a retrieval event. Retrieval only succeeds if the activation of a to-be-retrieved chunk is above a minimum threshold, which is a parameter in ACT-R.

The activation of a chunk is determined by several constraints. 
Let $C$ be the set of all chunks in declarative memory. The total activation of a chunk $i \in C$ equals

\begin{equation}\label{eq:1}
A_i = B_i + S_i + P_i + \epsilon,
\end{equation}

\noindent
where $B_i$ is the base-level or resting-state activation of the chunk $i$; the second summand $S_i$ represents the spreading activation that a chunk $i$ receives during a particular retrieval event; the third summand is a penalty for mismatches between a cue value $j$ and the value in the corresponding slot of chunk $i$; and finally, $\epsilon$ is noise that is logistically distributed, approximating a normal distribution, with location $0$ and scale ANS which is related to the variance of the distribution. It is generated at each new retrieval request.
The retrieval time $T_i$ of a chunk $i$ depends on its activation $A_i$ via $T_i = F \exp(-A_i)$, where $F$ is a scaling constant which we kept constant at $0.2$ here.

The scale parameter ANS of the logistic distribution from which $\epsilon$ is generated can be interpreted as implementing the \emph{intermittent deficiency} hypothesis, because higher values of ANS will tend to lead to more fluctuations in activation of a chunk and therefore higher rates of retrieval failure.\footnote{As an aside, note that \citeA{PatilEtAl2016} implemented intermittent deficiency using another source of noise in the model (utility noise). In future work, we will compare the relative change in quality of fit when intermittent deficiency is implemented in this way.} 
Increasing ANS leads to a larger influence of the random element on a chunk's activation, which represents the core idea of \emph{intermittent deficiency}: that there is not a constantly present damage to the processing system, but rather that the deficit occasionally interferes with parsing, leading to more errors.

The second summand in \eqref{eq:1}, representing the process of \emph{spreading activation} within the ACT-R framework, can be made more explicit for the goal buffer and for retrieval cues $j \in \{1, \ldots, J\}$ as 

\begin{equation}\label{eq:2}
  S_i = \sum_{j=1}^J W_j S_{ji}.
\end{equation}

\noindent
Here, $W_j = \frac{\text{GA}}{J}$, where GA is the \emph{goal activation} parameter and $S_{ji}$ is a value that increases for each matching retrieval cue. $S_{ji}$ reflects the association between the content of the goal buffer and the chunk $i$. 
The parameter GA determines the total amount of activation that can be allocated for all cues $j$ of the chunk in the goal buffer. It is a free parameter in ACT-R. This parameter, sometimes labelled the ``$W$ parameter'', has already been used to model individual differences in working memory capacity \cite{DailyEtAl2001}. 
Thus, it can be seen as one way (although by no means the only way) to implement the resource reduction hypothesis. 
The lower the GA value, the lower the difference in activation between the retrieval target and other chunks. This leads to more retrieval failures and lower differences in retrieval latency on average.

Finally, the hypothesis of \emph{slowed processing} can be mapped to the \emph{default action time} DAT in ACT-R. This defines the constant amount of time it takes a selected production rule to ``fire'', i.e.\ to start the actions specified in the action part of the rule. Higher values would lead to a higher delay in firing of production rules. Due to the longer decay in this case, retrieval may be slower and more retrieval failures may occur.

Next, we evaluate whether there is evidence consistent with the claims regarding slowed processing, intermittent deficiency, and resource reduction, when implemented using the parameters described above. 

\section{Simulations}

In this section we describe our modelling method and the procedure we use for fitting the model results to the empirical data from \citeA{CaplanEtAl2015}.

\subsection{Materials}

We used the data from 56 IWA and 46 matched controls published in \citeA{CaplanEtAl2015}. In this data-set, participants listened to recordings of sentences presented word-by-word; they paced themselves through the sentence, providing self-paced listening data. Participants processed 20 examples of 11 spoken sentence types and indicated which of two pictures corresponded to the meaning of each sentence. This yielded accuracy data for each sentence type. 

We chose two of the 11 sentence types for the current simulation: simple subject relatives (\textit{The woman who hugged the girl washed the boy}) vs.\ object relatives (\textit{The woman who the girl hugged washed the boy}), and subject relatives with a reflexive (\textit{The woman who hugged the girl washed herself}) vs.\ object relatives with a reflexive (\textit{The woman who the girl hugged washed herself}).
We chose relative clauses for two reasons. First, relative clauses have been very well-studied in psycholinguistics and serve as a typical example where processing difficulty is (arguably) experienced due to deviations in canonical word ordering \cite{jc92}.
Second, 
the \citeauthor{LewisVasishth2005} model already has productions defined for these constructions, so the relative clause data serve as a good test of the model as it currently stands.
The reflexive in the second sentence type adds an additional layer of complexity to the sentences. In the model, this is reflected by an additional retrieval process on the reflexive, where the antecedent is retrieved.

The \citeA{CaplanEtAl2015} dataset only provides accuracy data for the dependency between the embedded verb and its subject. We will address this problem in future studies where new data will be collected.

Lastly, since the production rules in the model were designed for modelling unimpaired processing, using them for IWA amounts to assuming that there is no damage to the parsing system per se, but rather that the processing problems in IWA are due to some subset of the cognitive constraints discussed earlier. 
This also implies that the IWA's parsing system is not engaged in heuristic processing, as has sometimes been claimed in the literature; see \citeA{PatilEtAl2016} for discussion on that point.

\begin{table*}[htpb!]
  \centering
  \begin{tabular}{llccccccc}
    ~  & ~        & GA & DAT & ANS & GA \& DAT & GA \& ANS & DAT \& ANS & GA \& DAT \& ANS\\
    \hline
    SR & control  & 19 & 24  & 18  & 18     & 11     & 16      & 10\\
    ~  & IWA      & 38 & 41  & 42  & 32     & 33     & 36      & 27\\
    \hline
    OR & control  & 21 & 26  & 36  & 21     & 20     & 25      & 20\\
    ~  & IWA      & 40 & 48  & 53  & 38     & 40     & 48      & 38\\
    \hline
  \end{tabular}
  \caption{Number of participants in \textbf{simple subject / object relatives} for which non-default parameter values were predicted, in the subject vs.\ object relative tasks, respectively; for goal activation (GA), default action time (DAT) and noise (ANS) parameters.}
  \label{table:normsettingsSO}
\end{table*}

\begin{table*}[htpb!]
  \centering
  \begin{tabular}{llccccccc}
    ~  & ~        & GA & DAT & ANS & GA \& DAT & GA \& ANS & DAT \& ANS & GA \& DAT \& ANS\\
    \hline
    SR & control  & 17 & 36  & 23  & 11        & 11        & 5          & 5\\
    ~  & IWA      & 40 & 46  & 42  & 36        & 35        & 31         & 31\\
    \hline
    OR & control  & 28 & 26  & 37  & 27        & 19        & 27         & 18\\
    ~  & IWA      & 51 & 48  & 51  & 44        & 46        & 41         & 39\\
    \hline
  \end{tabular}
  \caption{Number of participants in \textbf{subject / object relatives with reflexives} for which non-default parameter values were predicted, in the subject vs.\ object relative tasks, respectively; for goal activation (GA), default action time (DAT) and noise (ANS) parameters.}
  \label{table:normsettingsSOREF}
\end{table*}
\subsection{Method}

For the simulations, we refer to as the parameter space $\Pi$ the set of all vectors $(\text{GA}, \text{DAT}, \text{ANS})$ with $\text{GA},\ \text{DAT},\ \text{ANS} \in \mathbb{R}$.
For computational convenience, we chose a discretisation of $\Pi$ by defining a step-width and lower and upper boundaries for each parameter. In this discretised space $\Pi'$, we chose $\text{GA} \in \{0.2, 0.3, \ldots, 1.1\}$, $\text{DAT} \in \{0.05, 0.06, \ldots, 0.1\}$, and $\text{ANS} \in \{0.15, 0.2, \ldots, 0.45\}$.\footnote{The standard settings in the \citeA{LewisVasishth2005} model are $\text{GA} = 1,\ \text{DAT} = 0.05\ \text{(or 50 ms)},\ \text{and ANS} = 0.15$.} $\Pi'$ could be visualised as a three-dimensional grid of 420 dots, which are the elements $p' \in \Pi'$.


The default parameter values were included in $\Pi'$. This means that models that vary only one or two of the three parameters were included in the simulations. This is motivated by the results of \citeA{PatilEtAl2016}: there, the combined model varying both parameters (default action time (DAT) and utility noise) achieved the best fit to the data. Including all models allows us to do a similar investigation.

For all participants in the \citeA{CaplanEtAl2015} data-set, we calculated comprehension question response accuracies, averaged over all items of the subject / object relative clause and subject / object relative clause with reflexive conditions. For each $p' \in \Pi'$, we ran the model for 1000 iterations for the subject and object relative tasks.
From the model output, we determined whether the model made the correct attachment in each iteration, i.e.\ whether the correct noun was selected as subject of the embedded verb, and we calculated the accuracy in a simulation for a given parameter $p' \in \Pi'$ as the proportion of iterations where the model made the correct attachment. We counted a parsing failures, where the model did not create the target dependency, as an incorrect response.

The problem of finding the best fit for each subject can be phrased as follows: for all subjects, find the parameter vector that minimises the absolute distance between the model accuracy for that parameter vector and each subject's accuracy. Because there might not always be a unique $p'$ that solves this problem, the solution can be a set of parameter vectors.
If for any one participant multiple optimal parameters were calculated, we averaged each parameter value to obtain a unique parameter vector. This transforms the parameter estimates from the discretised space $\Pi'$ to the original parameter space $\Pi$.

\subsection{Results}

In this section we presents the results of the simulations and the fit to the data. First, we describe the general pattern of results reflected by the distribution of non-default parameter estimates per subject. Following that, we test whether tighter clustering occurs in controls.

\paragraph{Distribution of normal parameter values} 
Tables~\ref{table:normsettingsSO} and \ref{table:normsettingsSOREF} show the number of participants for which a non-default parameter value was predicted. 
By default values we mean the values $\text{GA} = 1,\ \text{DAT} = 0.05\ \text{(or 50 ms)},\ \text{and ANS} = 0.15$.
It is 
clear that, as expected, the number of subjects with non-default parameter values is always larger for IWA vs.\ controls, but controls show non-default values unexpectedly often.
In controls, the main difference between subject and object relatives is a clear increase in elevated noise values in object relatives for both simple subject / object relatives and those with reflexives. Perhaps surprisingly, in the reflexives condition (cf.\ Table~\ref{table:normsettingsSOREF}), controls display higher DAT in subject vs.\ object relatives.

For IWA in simple subject relatives, the single-parameter models are very similar, whereas in simple object relatives, most IWA (95\%) exhibit elevated noise values, while a far smaller proportion (71\%) showed reduced goal activation values. In the relatives with reflexives, IWA show the same pattern in subject and object relatives, with a high degree of non-default parameter estimates for each of the three parameters.

Overall, most IWA exhibit non-default parameter settings ANS and DAT. While in subject / object relatives with reflexives, a similar number of IWA shows elevated GA settings, we think this might be due to the similar model behaviours that non-default GA and ANS elicit. We address this point in the discussion below.

\paragraph{Cluster analysis} In order to investigate the
 predicted clustering of parameter estimates, we performed a cluster analysis on the data too see to which degree controls and IWA could be discriminated.
If our prediction is correct that, compared to IWA, clustering is tighter in controls, we expect that a higher proportion of the data should be correctly assigned to one of two clusters, one corresponding to controls, the other one corresponding to IWA. We chose hierarchical clustering to test this prediction.

 We combined the data for subject and object relatives into one respective data set, one for simple relatives, and one for relatives with reflexives.
 We calculated the dendrogram and cut the tree at 2, because we are only looking for the discrimination between controls and IWA. The results of this are shown in Table~\ref{table:hclustSO} and \ref{table:hclustSOREF}. In simple relatives (cf.\ Table~\ref{table:hclustSO}), the clustering is able to identify controls better than IWA, but the identification of IWA is better than chance (50\%). In relatives with reflexives (cf.\ Table~\ref{table:hclustSOREF}), clustering shows moderate but above chance discrimination ability in subject relatives. In object relatives with reflexives, controls are discriminated barely above chance, while there is an above chance proportion of misclassifications in IWA, demonstrating poor performance of the clustering there.
Discriminative ability might improve if all 11 constructions in \citeA{CaplanEtAl2015} were to be used; this will be investigated in future work.

\begin{table}
\begin{tabular}{ccccc}
 &\multicolumn{2}{c}{Subject relatives} & \multicolumn{2}{c}{Object relatives}\\
predicted group &    controls &  IWA & controls & IWA \\ 
         control      &  \textbf{34}             & 21   &        \textbf{42}   & 24\\
         IWA           & 12              & \textbf{35}    &         4    & \textbf{32}\\ 
 \hline        
         accuracy & 74\% & 63\% & 91\% & 57\%
\end{tabular}
\caption{Discrimination ability of hierarchical clustering on the combined data for \textbf{simple subject / object relative clauses}. Numbers in bold show the number of correctly clustered data points. The bottom row shows the percentage accuracy.}
  \label{table:hclustSO}
\end{table}

\begin{table}
  \begin{tabular}{ccccc}
 &\multicolumn{2}{c}{Subject relatives} & \multicolumn{2}{c}{Object relatives}\\
predicted group &    controls &  IWA & controls & IWA \\ 
         control      &  \textbf{31}             & 17   &        \textbf{27}   & 45\\
         IWA           & 15              & \textbf{39}    &         19    & \textbf{11}\\ 
 \hline        
         accuracy & 67\% & 70\% & 59\% & 20\%
\end{tabular}
\caption{Discrimination ability of hierarchical clustering on the combined data for \textbf{subject / object relative clauses with reflexives}. The numbers in bold are the correct classifications of controls/IWA. The bottom row shows the percentage accuracy.}
  \label{table:hclustSOREF}
\end{table}

\section{Discussion}

The simulations and cluster analysis above demonstrate overall tighter clustering in parameter estimates for controls, and more variance in IWA. This is evident from the clustering results in Tables~\ref{table:hclustSO} and \ref{table:hclustSOREF}.
These findings are consistent with the predictions of the small-scale study in \citeA{PatilEtAl2016}. However, there is considerable variability even in the parameter estimates for controls, more than expected based on the results of \citeA{PatilEtAl2016}.

The distribution of non-default parameter estimates (cf.\ Tables~\ref{table:normsettingsSO} and \ref{table:normsettingsSOREF})
suggest that all three hypotheses are possible explanations for the patterns in our simulation results: compared to controls, estimates for IWA tend to include higher default action times and activation noise scales, and lower goal activation. These effects generally appear to be more pronounced in object relatives vs.\ subject relatives. This means that all the three hypotheses can be considered viable candidate explanations. 
Overall, more IWA than controls display non-default parameter settings. Although there is evidence that many IWA are affected by all three impairments in our implementation, there are also many patients that show only one or two non-default parameter values. Again, this is more the case in object relatives than in subject relatives.

In general, there is evidence that all three deficits are plausible to some degree. However, IWA differ in the degree of the deficits, and they have a broader range of parameter values than controls.
Nevertheless, even the controls show a broad range of differences in parameter values, and even though these are not as variable as IWA, this suggests that some of the unimpaired controls can be seen as showing slowed processing, intermittent deficiencies, and resource reduction to some degree.   

There are several problems with the current modelling method. First, using the ACT-R framework with its multiple free parameters has the risk of overfitting. We plan to address this problem in three ways in future research. (1) Testing more constructions from the \citeA{CaplanEtAl2015} data-set might show whether the current estimates are unique to this kind of construction, or if they are generalisable. (2) We plan to create a new data-set analogous to Caplan's, using German as the test language. Once the English data-set has been analysed and the conclusions about the different candidate hypotheses have been tested on English, a crucial test of the conclusions will be cross-linguistic generalisability.
(3) We plan to investigate whether an approach as in \citeA{NicenboimVasishth2017StanCon}, using lognormal race models and mixture models, can be applied to our research question. 


Second, the use of accuracies as modelling measure has some drawbacks. Informally, in an accuracy value there is less information encoded than in, for example, reading or listening times. In future work, we will implement an approach modelling both accuracies and listening times. Also, counting each parsing failure as `wrong' might yield overly conservative accuracy values for the model; this will be addressed by assigning a random component into the calculation. This reflects more closely a participant who guesses if he/she did not fully comprehend the sentence.

Lastly, simulating the subject vs.\ object relative tasks separately yields the undesirable interpretation of participants' parameters varying across sentence types. While this is not totally implausible, estimating only one set of parameters for all sentence types would reduce the necessity of making additional theoretical assumptions on the underlying mechanisms, and allows for easier comparisons between different syntactic constructions.  We plan to do this in future work.

Although our method, as a proof of concept, showed that all three hypotheses are supported to some degree, it is worth investigating more thoroughly how different ACT-R mechanisms are influenced by changes in the three varied parameters in the present work. Implementing more of the constructions from \citeA{CaplanEtAl2015} will, for example, enable us to explore how the different hypotheses interact with each other in our implementation. 
More specifically, the decision to use the ANS parameter makes the assumption that the high noise levels for IWA influence all declarative memory retrieval processes, and thus the whole memory, not only the production system. Also, as both the GA and ANS parameters lead to higher failure rates, it will be worth investigating in future work whether a more focussed source of noise, such as utility noise, may be a better way to model intermittent deficiencies.

One possible way to delve deeper into identifying the sources of individual variability in IWA could be to investigate whether sub-clusters show up within the IWA parameter estimates.
For example, different IWA being grouped together by high noise values could be interpreted as these patients sharing a common source of their sentence processing deficit (in this hypothetical case, our implementation of intermittent deficiencies). We will address this question once we have simulated data for more constructions of the \citeA{CaplanEtAl2015} data-set.

\section{Acknowledgements}

Paul M\"{a}tzig was funded by the Studienstiftung des deutschen Volkes. 
This research was partly funded by the Volkswagen Foundation grant 89 953 to Shravan Vasishth.

\bibliographystyle{apacite}

\setlength{\bibleftmargin}{.125in}
\setlength{\bibindent}{-\bibleftmargin}

\bibliography{maetzigetal_iccm17}

\end{document}